\documentclass[10pt,twocolumn,letterpaper]{article}

\usepackage{cvpr}              
\definecolor{cvprblue}{rgb}{0.21,0.49,0.74}
\usepackage[pagebackref,breaklinks,colorlinks,allcolors=cvprblue]{hyperref}
\usepackage[table]{xcolor}
\usepackage{multirow}
\usepackage{makecell}
\usepackage{adjustbox}
\usepackage{booktabs}
\usepackage{pifont}
\newcommand{\cmark}{\ding{51}} 
\def\paperID{45594} 
\def\confName{CVPR}
\def\confYear{2026}

\title{Back to Point: Exploring Point-Language Models for Zero-Shot 3D Anomaly Detection}

\author{
Kaiqiang Li$^{1,2}$ \quad Gang Li$^{1,2}$\thanks{Co-corresponding authors.} \quad Mingle Zhou$^{1,2}$ \quad Min Li$^{1,2}$ \quad Delong Han$^{1,2}$ \quad Jin Wan$^{1,2}$\footnotemark[1]\\
$^{1}$Key Laboratory of Computing Power Network and Information Security, \\Ministry of Education, Shandong Computer Science Center (National Supercomputer Center in Jinan), \\Qilu University of Technology (Shandong Academy of Sciences), Jinan, China.\\
$^{2}$Shandong Provincial Key Laboratory of Computing Power Internet and Service Computing, \\Shandong Fundamental Research Center for Computer Science, Jinan, China.\\
\tt\small b1043125006@stu.qlu.edu.cn, \{lig, zhouml, limin, handl, wanj\}@qlu.edu.cn\\
}

\begin{document}
\maketitle
\begin{abstract}
Zero-shot (ZS) 3D anomaly detection is crucial for reliable industrial inspection, as it enables detecting and localizing defects without requiring any target-category training data. Existing approaches render 3D point clouds into 2D images and leverage pre-trained Vision-Language Models (VLMs) for anomaly detection. However, such strategies inevitably discard geometric details and exhibit limited sensitivity to local anomalies.  In this paper, we revisit intrinsic 3D representations and explore the potential of pre-trained Point-Language Models (PLMs) for ZS 3D anomaly detection. We propose BTP (Back To Point), a novel framework that effectively aligns 3D point cloud and textual embeddings. Specifically, BTP aligns multi-granularity patch features with textual representations for localized anomaly detection, while incorporating geometric descriptors to enhance sensitivity to structural anomalies. Furthermore, we introduce a joint representation learning strategy that leverages auxiliary point cloud data to improve robustness and enrich anomaly semantics. Extensive experiments on Real3D-AD and Anomaly-ShapeNet demonstrate that BTP achieves superior performance in ZS 3D anomaly detection. Code will be available at \href{https://github.com/wistful-8029/BTP-3DAD}{https://github.com/wistful-8029/BTP-3DAD}.
\end{abstract}

\section{Introduction}
\label{sec:intro}

\begin{figure}[t]
\centering
\captionsetup[subfigure]{justification=centering}
\subfloat[Existing VLM-based approaches]{
    \includegraphics[width=1\linewidth]{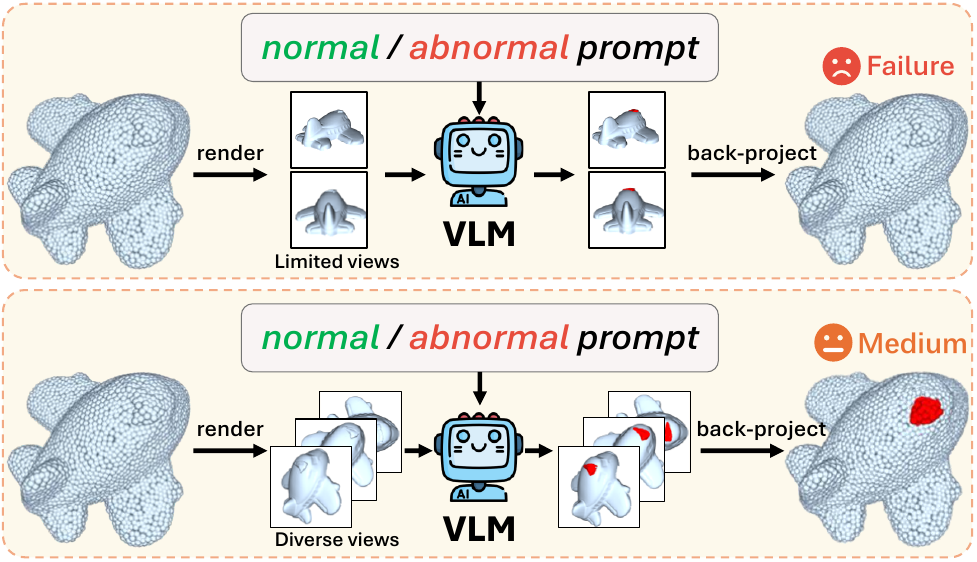}
    }
\hfill
\subfloat[Our proposed PLM-based approach]{
    \includegraphics[width=1\linewidth]{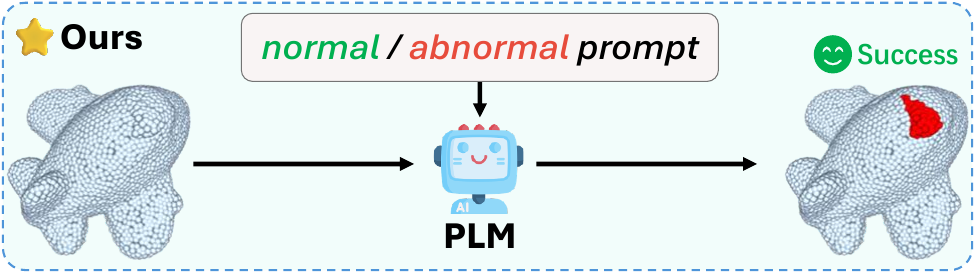}
    }
\caption{Comparison between VLM-based and PLM-based (ours) zero-shot 3D anomaly detection. (a): VLM-based approaches rely on multi-view rendering and back-projection, making their anomaly localization performance sensitive to the number and angles of rendered views. (b): Our proposed PLM-based approach directly processes point clouds, avoiding such view-dependent limitations and achieving more accurate 3D anomaly localization.}
\label{fig:motivation}
\end{figure}


3D anomaly detection is pivotal to industrial quality inspection, as it safeguards the safety, functionality, and reliability of products \cite{ye2025po3ad,you2022unified, wang2025multi, cao2024complementary, li2024towards, wang2025exploiting, zhou2024r3d, chen2023easynet, horwitz2023back, huo2023research,da20233d}. Existing unsupervised methods \cite{zhou2024r3d, li2024towards, cao2024complementary, horwitz2023back,chu2023shape,li2025multi} mainly rely on memory-bank retrieval or reconstruction-based paradigms, yet they struggle to generalize to unseen anomaly types. Moreover, practical deployment is often impeded by privacy restrictions and data acquisition limitations, which make it challenging to collect sufficiently diverse training samples in real-world scenarios \cite{khan20243dffl, lvone, lu2021data, bergmann2019mvtec,zhou2021novel, wang2021toward, zhu2025real,zhang2025zero}. Consequently, zero-shot (ZS) 3D anomaly detection has emerged as a promising paradigm, enabling defect detection and localization without anomalous samples, and effectively addressing the challenges of data scarcity and unseen anomaly generalization \cite{zhou2024pointad, wang2025towards,mao2025beyond,feng20213d}.


Recent advances in large-scale Vision-Language Models (VLMs) have significantly advanced ZS learning, resulting in remarkable progress in 2D anomaly detection. Mainstream approaches \cite{jeong2023winclip, zhou2024anomalyclip, ma2025aa, cao2024adaclip, NEURIPS2022_e43a3399,cao2025personalizing,xu2025towards,qu2025bayesian,chen2025distilled,mao2025beyond} leverage pre-trained VLMs (\textit{e.g.,} CLIP \cite{radford2021learning}) to align textual descriptions of normal and abnormal conditions with image features, thereby enabling anomaly detection without requiring any anomalous samples. Motivated by VLMs, 3D anomaly detection methods \cite{zhou2024pointad,cheng2025toward,li2025mcl} typically render 3D point clouds into multi-view 2D images, conduct anomaly detection in the 2D space, and then back-project the results onto the 3D point cloud, as illustrated in ~\cref{fig:motivation}. While these frameworks are consistent with 2D anomaly detection pipelines, they inevitably discard rich 3D geometric cues and show limited sensitivity to fine-grained structural anomalies. Moreover, their performance heavily depends on the number and angles of rendered views, making them vulnerable to view-selection bias. The repeated projection-back-projection process also introduces non-negligible computational overhead, further restricting scalability and efficiency.

Following the development of VLMs, the 3D domain has recently witnessed the rise of Point-Language Models (PLMs). ULIP~\cite{xue2023ulip}, the first framework to bridge VLMs with 3D point clouds, enabling unified representation learning across point clouds, images, and texts. By directly encoding 3D point clouds, ULIP preserves intrinsic geometric and structural properties, making it particularly suitable for ZS 3D anomaly detection, where capturing fine-grained spatial cues is essential. To this end, we explore the potential of PLMs for 3D anomaly detection and propose BTP (Back To Point)---a novel ZS 3D anomaly detection framework. Specifically, 3D anomalies often manifest as subtle structural deformations and local geometric variations. To capture the nuances, we design a Multi-Granularity Feature Embedding Module (MGFEM) that integrates multi-scale patch features, geometric descriptors, and global semantic representations into a shared text-aligned embedding space, capturing both fine-grained local deviations and holistic object semantics. Furthermore, to improve robustness across diverse anomaly patterns, we introduce a joint representation learning strategy that simultaneously optimizes global semantics, local patch-text alignment, and geometry-aware representations. This joint optimization allows BTP to learn complementary cues across multiple granularities, resulting in more reliable and generalizable ZS 3D anomaly detection. These designs collectively enable BTP to bridge point-language representations and achieve fine-grained, geometry-consistent ZS 3D anomaly detection.

 Our main contributions are summarized as follows:
\begin{itemize}
    \item To the best of our knowledge, we are the first to employ pre-trained PLMs for zero-shot 3D anomaly detection. Building upon this, we propose BTP, a novel framework that performs direct anomaly detection and localization on point clouds.
    \item We propose a multi-granularity feature embedding module that performs structure-aware alignment across patch-level, geometric, and global semantic features with textual embeddings, enabling fine-grained and geometry-consistent 3D anomaly localization.
    \item We design a joint representation learning strategy that jointly optimizes global semantic understanding, local geometry-text alignment, and fine-grained localization, resulting in more robust and discriminative anomaly representations.
    \item Extensive experiments on Real3D-AD and Anomaly-ShapeNet demonstrate that BTP achieves superior performance in ZS 3D anomaly detection.
\end{itemize}

\section{Related Work}
\label{sec:related}
\subsection{3D Anomaly Detection}
Due to the characteristics of 3D point cloud data and the scarcity of annotated samples, 3D anomaly detection poses greater challenges than its 2D counterpart. Existing approaches follow two paradigms: reconstruction-based \cite{li2024towards, zhou2024r3d, liu2023real3d} or contrastive-based methods \cite{cao2024complementary}. IMRNet \cite{li2024towards} employs masked reconstruction for anomaly localization, while R3D-AD \cite{zhou2024r3d} leverages diffusion for point cloud reconstruction. Reg3DAD \cite{liu2023real3d} builds a memory bank of normal samples using a pretrained 3D feature extractor, and CPMF \cite{cao2024complementary} fuses handcrafted 3D and 2D features via multi-view projections.
Despite these advances, ZS 3D anomaly detection remains in its infancy. The key challenge lies in learning transferable 3D representations that generalize to unseen defects without anomalous supervision \cite{zhu2022multi}. Studies such as PointAD \cite{zhou2024pointad} and MVP \cite{cheng2025toward} project point clouds into multi-view 2D images and employ pretrained VLMs for detection. However, such projection-based pipelines discard geometric structures and are sensitive to view selection.
Meanwhile, PLM-based extensions (e.g., PLANE \cite{wang2025exploiting}) adapt pretrained Point-Language Models to 3D anomaly detection via learnable prompts and self-supervised training, but still rely on target-category data for category-specific training/adaptation.
In contrast, BTP performs ZS 3D anomaly detection directly in 3D space without target-category data, eliminating 3D-to-2D conversion, preserving geometric fidelity, and enabling end-to-end, geometry-consistent localization.

\begin{figure*}[t]
    \centering
    \includegraphics[width=1\linewidth]{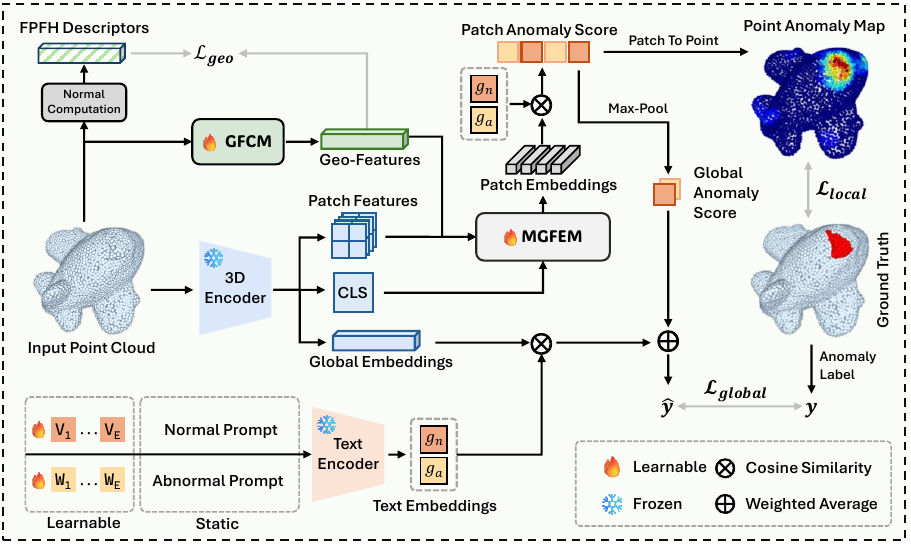}
    \caption{Overview of the BTP. The input point cloud is first processed by the 3D encoder and the Geometric Feature Creation Module (GFCM) to extract implicit semantic features and explicit geometric descriptors, respectively. The semantic features include patch features, global embeddings, and the CLS token. To enable anomaly localization, patch features, the CLS token, and geometric descriptors are feed into the Multi-Granularity Feature Embedding Module (MGFEM) to generate patch-level embeddings, which are compared with text embeddings for point-level anomaly detection. Global embeddings are aligned with text embeddings for object-level anomaly detection. A joint loss integrates geometric, local, and global supervision signals to jointly optimize GFCM, MGFEM, and the learnable text prompts.}
    \label{fig:overview}
\end{figure*}

\subsection{3D Feature Extraction}
The evolution of 3D point cloud feature extraction has progressed from traditional hand-crafted designs to deep learning-driven approaches. Early methods mainly relied on geometric descriptors, such as FPFH~\cite{rusu2009fast}, SHOT~\cite{salti2014shot}, and Spin Images~\cite{johnson2002using}. While these descriptors encode local geometric structures, their representational capacity is limited, making it difficult to capture high-level semantic information. Moreover, they often perform poorly under noise and sparse point distributions~\cite{guo2020deep,guo2016comprehensive}.
With the advancement of deep learning, neural network-based methods have become mainstream. PointNet~\cite{qi2017pointnet} pioneered direct processing of unordered point sets by employing shared multilayer perceptrons to extract point-wise features. Building on this foundation, PointNet++~\cite{qi2017pointnet++} introduced hierarchical sampling and local feature aggregation to better capture local structures. DGCNN~\cite{wang2019dynamic} further enhanced the representation of local geometric relationships by constructing dynamic graphs. These methods demonstrated strong performance in 3D recognition and detection tasks, yet still face challenges in modeling long-range dependencies and processing large-scale point clouds.
Inspired by the success of Transformer~\cite{vaswani2017attention} architectures in natural language processing and computer vision, researchers have introduced them into 3D point cloud feature learning. Point-BERT~\cite{yu2022point}, a representative approach, leverages BERT~\cite{devlin2019bert} and adopts a mask-and-reconstruct self-supervised pretraining paradigm. This paradigm enables the model to capture rich contextual and semantic representations from large-scale unlabeled point cloud data. Compared with hand-crafted or shallow learning-based methods, Point-BERT excels at modeling long-range dependencies and bridges the gap between 3D and vision-language modalities, making it a widely adopted point cloud encoder in PLMs~\cite{xue2023ulip,xue2024ulip,liu2023openshape}.
Building upon this trend, PLMs extend 3D representation learning to jointly align geometric and textual embeddings, unifying multi-modal features for ZS 3D understanding.

\section{Methods}

\subsection{A Review of ULIP}
ULIP, a representative PLM, is built upon the core idea of unifying the feature space of a 3D encoder with the pre-aligned vision-language embedding space, thereby enabling a joint representation of points, language, and vision. Specifically, given a point cloud-text-image triplet, ULIP leverages a pre-trained VLM as the alignment reference and trains a 3D encoder such that point cloud features can be embedded into the shared semantic space of images and text. For the point cloud classification task, given a point cloud 
$P$ and a set of target classes $C$, the probability distribution on a specific category $c$ can be expressed as:
\begin{equation}
\small
P(y = c \mid P) = 
\frac{\exp \big( \cos(g_c, f_P) / \tau \big)}
{\sum_{c \in C} \exp \big( \cos(g_{c}, f_P) / \tau \big)} ,
\end{equation}
where $\cos(\cdot, \cdot)$ and $\tau$ denote the cosine similarity and temperature parameter, respectively. $f_P$ represents the feature embedding of the point cloud $P$, and $g_c$ denotes the text prompt embedding of class $c$. The text prompt follows the template \texttt{"a point cloud model of a [c]"}, where $c \in C$.

\subsection{Overview}

In this paper, we propose BTP, a ZS 3D anomaly detection framework that directly extracts multi-granularity representations from point clouds to perform fine-grained anomaly reasoning. As illustrated in ~\cref{fig:overview}, BTP first leverages a pre-trained ULIP encoder to obtain a global point cloud embedding, which supports object-level anomaly detection by measuring semantic consistency between the point cloud and textual descriptions.
To better capture structural anomalies, we introduce a geometry-aware branch that incorporates FPFH descriptors to impose explicit geometric constraints on the learned features. For point-level anomaly detection, we fuse intermediate patch features, the CLS token, and geometry-aware features extracted from the ULIP encoder, and project them into a shared embedding space for unified cross-modal alignment. Finally, anomaly scores are computed by measuring the similarity between the fused point-cloud features and textual embeddings, enabling multi-granularity anomaly detection from global object-level reasoning down to individual points.

\subsection{Point Cloud Feature Extraction}
\textbf{Patch-level Feature Exploitation.} ULIP was originally designed to achieve unified representations across point clouds, images, and language, and its typical usage relies on the global embedding from the final encoder layer for point cloud classification. However, solely depending on the global embedding, while effective for capturing object-level semantics, is insufficient for precise localization of local anomalies. As a result, when directly applied to anomaly detection, ULIP tends to lack sensitivity to fine-grained structural variations. To address this limitation, we extend the feature utilization strategy of ULIP in BTP. Specifically, in addition to retaining the final global embedding, we extract patch-level representations from multiple intermediate layers of the encoder. Empirical results demonstrate that these patch-level representations capture geometric and semantic information at different levels of abstraction. By appropriately integrating them, our model achieves enhanced sensitivity to local structural variations and exhibits superior capability in fine-grained anomaly localization. More details of the implementation are provided in ~\cref{sec:imp_details}.

\noindent{\textbf{Geometric Feature Creation Module.} The Fast Point Feature Histogram (FPFH) is a classical handcrafted descriptor for point cloud processing, effectively characterizing local geometric relationships. It has been extensively employed in point cloud matching, registration, and anomaly detection tasks. However, as a non-learnable handcrafted descriptor, FPFH cannot be jointly optimized with network parameters in an end-to-end training pipeline, and its representational capacity is inherently limited by a fixed feature space. To overcome these limitations, we introduce a PointNet-based Geometric Feature Creation Module (GFCM) as a learnable alternative to FPFH. Specifically, given an input point cloud $P$ and its patch index set, we first sample neighborhood points according to the indices to obtain local point sets for each patch. Then, each point within a patch is processed through multiple layers of convolution, normalization, and non-linear activation to extract point-wise features. Subsequently, a max-pooling operation is applied within each patch to aggregate these features into a patch-level geometric descriptor. Finally, a fully connected layer projects the descriptor into an embedding space aligned with the dimensionality of the text embeddings, yielding the final geometric feature:}

\begin{equation}
\small
\mathbf{f}_i = \phi\!\left( \max_{j=1,\ldots,M} \mathrm{MLP}(\mathbf{p}_{ij}) \right),
\quad i = 1,\ldots,G,
\end{equation}
\noindent
where $\mathbf{p}_{ij}$ denotes the $j$-th point in the $i$-th patch, $\mathrm{MLP}(\cdot)$ is a shared point-wise multilayer perceptron, and $\max(\cdot)$ performs max-pooling across all points within each patch. The function $\phi(\cdot)$ denotes a nonlinear mapping implemented by fully connected layers. The resulting $\mathbf{f}_i \in \mathbb{R}^d$ represents the geometric feature of the $i$-th patch, where $d$ is the embedding dimension.

\subsection{Multi-Granularity Embedding Construction}
To achieve fine-grained anomaly localization, we propose the Multi-Granularity Feature Embedding Module (MGFEM), which fuses multi-layer intermediate semantic features of point clouds, local geometric descriptors, and the global CLS token at the patch level, thereby producing structure-aware patch representations. 
Formally, given a point cloud $P$, its geometric and semantic features are defined as:
\begin{equation}
\small
\mathbf{F}_{\mathrm{geo}} = \Phi_{\mathrm{GFCM}}(P), \quad 
\mathbf{H}^{(l)} = \Phi_{\mathrm{enc}}^{(l)}(P), \; l = 1, \ldots, L,
\end{equation}
where $\Phi_{\mathrm{GFCM}}(\cdot)$ denotes the geometric feature creation module that outputs $\mathbf{F}_{\mathrm{geo}} \in \mathbb{R}^{N \times d_f}$, and $\Phi_{\mathrm{enc}}^{(l)}(\cdot)$ denotes the $l$-th layer of the point cloud encoder, producing $\mathbf{H}^{(l)} \in \mathbb{R}^{N \times d_l}$. 
Here, $N$ denotes the number of patches, $d_f$ and $d_l$ denote the geometric and semantic feature dimensions, respectively, and $L$ is the total number of layers.

Our objective is to jointly model $\mathbf{F}_{\mathrm{geo}}$, $\mathbf{H}^{(l)}$, and $\mathbf{h}_{\mathrm{CLS}}$ to obtain patch representations $\mathbf{Z} \in \mathbb{R}^{N \times D}$ aligned with the dimensionality of the text embedding space.

The MGFEM integrates three types of information:  
(1) multi-layer intermediate semantic features from the PointBERT encoder, $\{\mathbf{H}^{(l)}\}_{l = 1}^{L}$;  
(2) local geometric features $\mathbf{F}_{\mathrm{geo}}$ extracted by the geometric feature creation module; and  
(3) the CLS token representation $\mathbf{h}_{\mathrm{CLS}}$.

\begin{table*}[t]
\centering
\caption{Comparison of object-level(O.) and point-level(L.) AUROC on Real3D-AD. All values are reported in percentage, where higher is better. The \colorbox{red!25}{\textbf{best}}, \colorbox{orange!25}{{second-best}}, and \colorbox{cyan!15}{zero-shot} results are highlighted. $\dagger$ indicates results taken from~\cite{zhou2024pointad}.}
\label{table:main}
\renewcommand{\arraystretch}{1.0}
\begin{adjustbox}{max width=\textwidth}
\begin{tabular}{c ccccccccccccc|c}
\toprule
Metric & Methods & airplane & car & candybar & chicken & diamond & duck & fish & gemstone & seahorse & shell & starfish & toffees & Mean \\
\midrule
\multirow{10}{*}{O.}

& CPMF \cite{cao2024complementary}
& \colorbox{red!25}{\textbf{70.1}} & 55.2 & 55.1 & 50.4 & 52.3 & 58.2 & 55.8 & 58.9 & \colorbox{orange!25}{{72.9}} & 65.3 & \colorbox{orange!25}{{70.0}} & 39.0 & 58.6 \\
& BTF \cite{horwitz2023back}
& 45.4 & \colorbox{orange!25}{{56.2}} & 54.5 & 47.8 & 50.2 & 60.5 & 39.8 & 41.0 & 63.4 & 62.7 & 43.5 & 47.3 & 51.0 \\

& M3DM\textsubscript{MAE} \cite{wang2023multimodal}
& 38.2 & 46.4 & 48.4 & \colorbox{orange!25}{{54.2}} & 66.6 & 45.1 & 54.2 & 61.7 & 66.1 & 64.1 & 47.5 & 62.3 & 54.6 \\

& M3DM\textsubscript{BERT} \cite{wang2023multimodal}
& 39.8 & 53.9 & \colorbox{orange!25}{{58.0}} & 53.9 & 66.8 & 50.3 & \colorbox{orange!25}{{63.0}} & 57.9 & 56.6 & 63.8 & 49.6 & 63.6 & 56.4 \\

& PatchCore\textsubscript{FPFH} \cite{roth2022towards}
& 57.4 & 52.8 & 45.8 & 54.0 & 56.2 & 52.0 & 46.6 & 42.8 & 56.4 & 57.2 & 66.0 & 50.4 & 53.1 \\

& PatchCore\textsubscript{MAE} \cite{roth2022towards}
& 59.8 & 42.4 & 45.2 & 51.4 & 65.7 & 48.9 & 47.8 & 49.8 & 52.7 & 54.8 & 45.5 & 42.2 & 50.5 \\

& \colorbox{cyan!15}{PointCLIPV2} \textdagger \cite{zhu2023pointclip}
& 49.9 & 41.7 & 44.8 & 46.0 & 51.5 & \colorbox{red!25}{\textbf{68.4}} & 60.9 & 69.6 & 53.1 & 41.9 & 31.1 & \colorbox{red!25}{\textbf{78.6}} & 53.1 \\
& \colorbox{cyan!15}{AnomalyCLIP} \textdagger \cite{zhou2024anomalyclip}
& \colorbox{orange!25}{{61.7}} & 51.2 & 49.7 & \colorbox{red!25}{\textbf{57.9}} & 65.0 & 56.2 & 56.4 & 49.1 & 56.5 & 53.1 & 54.8 & 51.0 & 55.2 \\

& \colorbox{cyan!15}{PointAD} \textdagger \cite{zhou2024pointad}
& 60.9 & \colorbox{red!25}{\textbf{73.9}} & \colorbox{red!25}{\textbf{74.1}} & 52.0 & \colorbox{red!25}{\textbf{99.2}} & 60.1 & \colorbox{red!25}{\textbf{74.3}} & \colorbox{red!25}{\textbf{87.3}} & \colorbox{red!25}{\textbf{76.9}} & \colorbox{red!25}{\textbf{89.5}} & \colorbox{red!25}{\textbf{80.9}} & 69.0 & \colorbox{red!25}{\textbf{74.8}} \\

& \colorbox{cyan!15}{\textbf{\texttt{BTP(Ours)}}}
& 42.9 & 55.9 & 56.3 & 48.5 & \colorbox{orange!25}{{94.4}} & \colorbox{orange!25}{{66.7}} & 59.6 & \colorbox{orange!25}{{71.7}} & 53.5 & \colorbox{orange!25}{{73.6}} & 38.3 & \colorbox{orange!25}{{74.9}} & \colorbox{orange!25}{{61.4}} \\

\midrule
\multirow{10}{*}{L.}
& CPMF \cite{cao2024complementary} 
& 61.8 & \colorbox{orange!25}{{83.6}} & \colorbox{orange!25}{{73.4}} & 55.9 & 75.3 & \colorbox{orange!25}{{71.9}} & \colorbox{red!25}{\textbf{98.8}} & 44.9 & \colorbox{red!25}{\textbf{96.2}} & 72.5 & \colorbox{orange!25}{{80.0}} & \colorbox{red!25}{\textbf{95.9}} & \colorbox{orange!25}{{75.9}} \\
& BTF \cite{horwitz2023back}
& 57.0 & 48.0 & 53.9 & 49.5 & 49.6 & 56.3 & 47.6 & 48.8 & 58.1 & 54.8 & 51.4 & 48.1 & 51.9 \\

& M3DM\textsubscript{MAE} \cite{wang2023multimodal}
& 49.7 & 54.2 & 63.7 & 62.8 & 61.3 & 62.7 & 55.4 & 64.9 & 55.6 & 70.1 & 53.0 & 64.2 & 59.8 \\

& M3DM\textsubscript{BERT} \cite{wang2023multimodal}
& 46.2 & 51.7 & 62.6 & 61.8 & 58.5 & 58.8 & 53.9 & 62.6 & 55.5 & 67.9 & 50.5 & 64.6 & 57.9 \\

& PatchCore\textsubscript{FPFH} \cite{roth2022towards}
& 53.7 & 52.5 & 57.0 & 62.2 & \colorbox{orange!25}{{88.0}} & 71.2 & 74.2 & 74.0 & 50.5 & 49.1 & 49.5 & 67.8 & 62.5 \\
& PatchCore\textsubscript{MAE} \cite{roth2022towards}
& 55.7 & 49.7 & 57.2 & 60.3 & 58.7 & 49.4 & 60.0 & 44.5 & 55.3 & 66.3 & 53.6 & 61.5 & 55.1 \\
& \colorbox{cyan!15}{PointCLIPV2}\textdagger \cite{zhu2023pointclip}
& 45.1 & 56.3 & 55.2 & 46.5 & 52.6 & 62.6 & 64.9 & 51.5 & 48.3 & 58.1 & 40.6 & 53.6 & 52.9 \\
& \colorbox{cyan!15}{AnomalyCLIP}\textdagger \cite{zhou2024anomalyclip}
& 51.1 & 48.8 & 51.7 & 50.0 & 55.2 & 48.9 & 46.5 & 48.9 & 49.2 & 50.6 & 51.0 & 51.1 & 50.3 \\
& \colorbox{cyan!15}{PointAD}\textdagger \cite{zhou2024pointad}
& \colorbox{orange!25}{{67.2}} & 72.3 & 71.3 & \colorbox{orange!25}{{67.7}} & 87.7 & 51.0 & 80.0 & \colorbox{orange!25}{{80.2}} & \colorbox{orange!25}{{74.8}} & \colorbox{orange!25}{{77.8}} & \colorbox{red!25}{\textbf{81.4}} & 70.0 & 73.5 \\
& \textbf{\texttt{\colorbox{cyan!15}{BTP(Ours)}}}
& \colorbox{red!25}{\textbf{75.9}} & \colorbox{red!25}{\textbf{91.6}} & \colorbox{red!25}{\textbf{77.7}} & \colorbox{red!25}{\textbf{85.0}} & \colorbox{red!25}{\textbf{97.9}} & \colorbox{red!25}{\textbf{90.9}} & \colorbox{orange!25}{{90.3}} & \colorbox{red!25}{\textbf{87.2}} & 68.8 & \colorbox{red!25}{\textbf{81.3}} & 74.5 & \colorbox{orange!25}{{93.2}} & \colorbox{red!25}{\textbf{84.5}} \\

\bottomrule
\end{tabular}
\end{adjustbox}
\end{table*}

Specifically, we first project each modality into a unified embedding space: 
$\mathbf{S}^{(l)} = \phi_s(\mathbf{H}^{(l)})$, 
$\mathbf{G} = \phi_g(\mathbf{F}_{\mathrm{geo}})$, and 
$\mathbf{C} = \phi_c(\mathbf{h}_{\mathrm{CLS}})$. 
Here, $\phi_s(\cdot)$, $\phi_g(\cdot)$, and $\phi_c(\cdot)$ denote learnable projection functions implemented by linear layers, 
and $\alpha_l$ are softmax-normalized weights assigned to each semantic layer. We then aggregate these representations through a fusion layer to obtain the final multi-granularity embedding:

\begin{equation}
\small
\mathbf{Z} = \phi_f\!\Big(
\big[\sum_{l=1}^{L} \alpha_l \mathbf{S}^{(l)} 
\Vert \mathbf{G} \Vert \mathbf{C}\big]
\Big),
\quad \mathbf{Z} \in \mathbb{R}^{N \times D}.
\end{equation}

The resulting $\mathbf{Z}$ integrates local geometric cues, hierarchical semantic features, and global structural priors, 
yielding structure-aware patch representations that facilitate fine-grained anomaly localization.

\subsection{Hybrid Learnable Prompt}
To adapt textual representations to the 3D anomaly detection scenario, we introduce a hybrid prompt learner that combines learnable context tokens with fixed template phrases. Specifically, a small number of learnable context tokens are defined and jointly optimized with the rest of the network, and are inserted between class-specific prefix and suffix tokens derived from manually designed templates such as \textit{“normal object”} and \textit{“defective object”}. In this way, the hybrid prompts preserve semantic priors from natural language while adaptively aligning with dataset-specific distributions. 
The resulting learnable prompts are encoded by ULIP to produce positive (normal) and negative (anomalous) textual embeddings. These embeddings are subsequently used to compute similarity with point-cloud features for ZS anomaly detection.

\subsection{Joint Representation Learning }
To achieve fine-grained anomaly localization and global discrimination capability while enhancing the structural awareness of geometric features, we introduce multi-level supervision signals during training. The overall loss function consists of point-level supervision $\mathcal{L}_{\text{local}}$, global supervision $\mathcal{L}_{\text{global}}$, and geometric supervision $\mathcal{L}_{\text{geo}}$, which provide complementary information at different levels and thereby improve the robustness and discriminative power of the model.

For point-level supervision, we jointly employ the focal loss and the dice loss. Focal loss adaptively adjusts the weighting of positive and negative samples to alleviate the severe imbalance between normal and anomalous points, while dice loss optimizes the overlap between the predicted masks and the ground-truth labels, improving the consistency of anomaly region coverage. The combination ensures both attention to hard-to-classify anomalies and improved overall region quality, and can be formulated as:
\begin{equation}
\mathcal{L}_{\text{local}} = \mathcal{L}_{\text{focal}} + \mathcal{L}_{\text{dice}}.
\end{equation}

To discriminate between normal and anomalous objects, 
we fuse point- and patch-level predictions and optimize a binary cross-entropy loss:
\begin{equation}
\small
\hat{y} = \alpha \hat{y}_{\text{pt}} + (1-\alpha)\hat{y}_{\text{patch}}, \quad
\mathcal{L}_{\text{global}} = \text{BCE}(\hat{y}, y),
\end{equation}
where $\alpha$ is a balancing coefficient.

Additionally, to strengthen the geometric feature creation module in modeling local structures, we introduce a geometric supervision loss. Concretely, the learned geometric features are aligned with FPFH descriptors via a contrastive objective, enabling the model to explicitly absorb handcrafted geometric priors and to become more sensitive to fine-grained structural discrepancies. A generic formulation is:
\begin{equation}
\mathcal{L}_{\text{geo}} = \frac{1}{N}\sum_{i=1}^{N} \, \ell_{\text{ctr}}\!\big(\,\phi_{\text{geo}}(\mathbf{f}^{(i)}_{\text{geo}}), \, \hat{\mathbf{f}}^{(i)}_{\text{FPFH}} \big),
\end{equation}
where $\phi_{\text{geo}}(\cdot)$ denotes the projection of learned geometric features, $\hat{\mathbf{f}}^{(i)}_{\text{FPFH}}$ is the FPFH descriptor of the $i$-th patch, and $\ell_{\text{ctr}}(\cdot,\cdot)$ is a contrastive loss (\textit{e.g.}, cosine-based InfoNCE).

By integrating the above three types of supervision, the final optimization objective is defined as:
\begin{equation}
\mathcal{L} = \mathcal{L}_{\text{local}} + \lambda_1 \mathcal{L}_{\text{global}} + \lambda_2 \mathcal{L}_{\text{geo}},
\end{equation}
where $\lambda_1$ and $\lambda_2$ are balancing coefficients, set to $0.5$ and $0.1$ in our experiments.

\section{Experiment}
\label{sec:exp}

\subsection{Datasets and Evaluation Metrics}

\textbf{Real3D-AD.} The Real3D-AD \cite{liu2023real3d} dataset is a real-world industrial 3D anomaly detection dataset covering 12 object categories, with 1,254 samples including normal and defective instances.

\noindent{\textbf{Anomaly-ShapeNet.} The Anomaly-ShapeNet \cite{li2024towards} dataset is a synthetic 3D anomaly detection dataset constructed from ShapeNetCoreV2 \cite{chang2015shapenet}, covering 40 categories, with more than 1,600 samples including both normal and defective instances.}

\noindent{\textbf{Evaluation Metrics.} 
We evaluate both object-level and point-level performance. 
For object-level detection, we report the Area Under the Receiver Operating Characteristic Curve (AUROC), Average Precision (AP), and maximum F1 score (F1), 
measuring the binary classification outcomes derived from the global anomaly score. 
For point-level localization, we use AUROC, AP, and Area Under the Per-Region Overlap curve (AU-PRO), 
which jointly reflect coverage of anomalous regions and control of false positives across thresholds. 
In addition to overall averages, we also report category-wise results for fair comparison.}

\subsection{Implementation Details}
\label{sec:imp_details}
Input point clouds are uniformly downsampled to $2{,}048$ points via farthest point sampling (FPS). We adopt the public ULIP2 \cite{xue2024ulip} as the 3D and text encoder. The length of the learnable text prompt is set to $4$. For anomaly localization, we extract intermediate features from layers ${2,5,8,11}$ of the 3D encoder to form multi-layer semantic representations, following the feature extraction strategy in \cite{wang2025exploiting}. We use the AdamW optimizer with parameter grouping for differential learning rates and weight decay. The learning rate schedule is a 10\% linear warmup followed by cosine annealing, with a minimum learning rate of $1\times 10^{-6}$; global gradient clipping is applied to stabilize training. All experiments were conducted on a single NVIDIA RTX 4090 24GB GPU using PyTorch-2.1.0. Unless otherwise specified, reported results are the average over 10 independent runs with different random seeds.

\begin{table*}[htbp]
\centering
\caption{Cross-category anomaly detection results (P-AUROC, \%) on Real3D-AD, 
where each row corresponds to training on a single category and testing across all categories. 
The last column reports the average performance of our proposed BTP method under each training setting.}
\label{table:train_real3d}
\renewcommand{\arraystretch}{1.05}
\footnotesize
\begin{adjustbox}{max width=\textwidth}
\begin{tabular}{ccccccccccccc|c}
\toprule
train\textbackslash test & airplane & candybar & car & chicken & diamond & duck & fish & gemstone & seahorse & shell & starfish & toffees & Mean\\
\midrule
airplane  & -   & 82.2 & 77.3 & 76.3 & 94.7 & 85.5 & 72.7 & 86.7 & 64.6 & 59.0 & 75.7 & 78.8 & 77.6\\
candybar  & 77.9 & -   & 63.1 & 82.3 & 86.1 & 86.4 & 84.6 & 81.2 & 66.2 & 82.2 & 58.9 & 90.7 & 78.1\\
car       & 76.1 & 88.8 & -   & 82.9 & 97.0 & 87.9 & 82.0 & 84.7 & 70.8 & 80.2 & 74.1 & 93.7 & 83.5\\
chicken   & 74.3 & 92.3 & 71.4 & -   & 97.2 & 92.0 & 85.0 & 89.6 & 65.6 & 77.3 & 72.5 & 93.3 & 82.8\\
diamond   & 75.8 & 86.6 & 76.1 & 83.9 & -   & 88.1 & 90.6 & 92.8 & 75.1 & 83.9 & 83.7 & 91.2 & \colorbox{orange!25}{84.3}\\
duck      & 73.2 & 86.0 & 83.1 & 82.8 & 97.1 & -   & 91.3 & 86.0 & 63.9 & 74.7 & 68.3 & 94.4 & 81.9\\
fish      & 68.2 & 86.9 & 68.6 & 83.6 & 88.5 & 92.9 & -   & 76.4 & 70.7 & 86.4 & 80.6 & 95.1 & 81.6\\
gemstone  & 77.2 & 93.7 & 83.9 & 87.2 & 99.5 & 92.8 & 93.8 & -   & 69.9 & 86.4 & 76.8 & 92.7 & \colorbox{red!25}{\textbf{86.7}}\\
seahorse  & 68.0 & 80.9 & 64.6 & 79.6 & 89.8 & 80.7 & 94.8 & 79.1 & -   & 77.5 & 82.1 & 87.0 & 80.4\\
shell     & 67.0 & 91.5 & 77.1 & 84.6 & 96.7 & 86.0 & 89.1 & 92.9 & 70.6 & -   & 67.2 & 91.5 & 83.1\\
starfish  & 73.3 & 76.1 & 69.9 & 76.6 & 98.1 & 89.8 & 91.8 & 89.0 & 76.7 & 80.8 & -   & 93.2 & 83.2\\
toffees   & 67.0 & 79.0 & 56.5 & 79.7 & 85.2 & 85.1 & 87.4 & 69.4 & 63.3 & 76.4 & 59.1 & -   & 73.5\\
\bottomrule
\end{tabular}
\end{adjustbox}
\end{table*}

\begin{table*}[h]
\centering
\caption{More comprehensive results on the Real3D-AD and the Anomaly-ShapeNet datasets. }
\renewcommand{\arraystretch}{1.05}
\label{tab:all_results}
\setlength{\tabcolsep}{3pt}
\begin{adjustbox}{max width=1\textwidth}
\begin{tabular}{c|cccccc|cccccc}
\toprule
\multirow{2}{*}{Methods} & \multicolumn{6}{c|}{Real3D-AD} & \multicolumn{6}{c}{Anomaly-ShapeNet} \\
\cmidrule(lr){2-7} \cmidrule(lr){8-13}
 & O-AUROC & O-AP & O-F1 & P-PRO & P-AUROC & P-AP & O-AUROC & O-AP & O-F1 & P-PRO & P-AUROC & P-AP \\
\midrule
BTF \cite{horwitz2023back}  & 51.0 & 61.0 & 70.0 & 18.6 & 51.9 & 2.2 & 51.0 & 58.3 & 72.2 & 15.3 & 51.0 & 1.8 \\
M3DM\textsubscript{MAE} \cite{wang2023multimodal}  & 54.6 & 54.7 & 68.4 & 27.6 & 59.8 & 2.3 & 55.1 & 62.0 & \colorbox{orange!25}{{73.1}} & \colorbox{orange!25}{{26.6}} & \colorbox{orange!25}{{65.1}} & 2.9 \\
M3DM\textsubscript{BERT} \cite{wang2023multimodal}  & 56.4 & 55.4 & 66.9 & 27.1 & 57.9 & 2.5 & 55.5 & 62.0 & 71.7 & 25.5 & 63.3 & 2.6 \\
Patchcore\textsubscript{MAE} \cite{roth2022towards}  & 50.5 & 60.2 & 68.7 & 29.7 & 55.1 & 5.7 & 54.7 & 61.8 & 72.5 & 22.1 & 64.5 & 2.8 \\
Patchcore\textsubscript{BERT} \cite{roth2022towards}  & 54.7 & 56.4 & 67.9 & 26.7 & 61.4 & 3.8 & 56.7 & \colorbox{orange!25}{{63.3}} & 71.3 & 22.1 & 64.8 & \colorbox{orange!25}{{3.3}} \\
Reg3DAD \cite{liu2023real3d}  & 69.0 & \colorbox{orange!25}{{70.9}} & \colorbox{red!25}{\textbf{73.1}} & \colorbox{orange!25}{{30.6}} & 69.5 & \colorbox{orange!25}{{11.5}} & 52.7 & 60.4 & 72.0 & 21.9 & 63.2 & 2.4 \\
R3DAD \cite{zhou2024r3d}  & \colorbox{orange!25}{{73.4}} & - & - & - & - & - & \colorbox{red!25}{\textbf{74.9}} & - & - & - & - & - \\
IMR \cite{li2024towards}  & 72.5 & - & - & - & - & - & \colorbox{orange!25}{{65.9}} & - & - & - & 65.0 & - \\
\colorbox{cyan!15}{PointCLIPV2} \cite{zhu2023pointclip} & 53.1 & 58.1 & - & - & 52.9 & - & - & - & - & - & - & - \\
\colorbox{cyan!15}{AnomalyCLIP} \cite{zhou2024anomalyclip} & 55.2 & 57.1 & - & - & 50.3 & - & - & - & - & - & - & - \\
\colorbox{cyan!15}{PointAD} \cite{zhou2024pointad}  & \colorbox{red!25}{\textbf{74.8}} & \colorbox{red!25}{\textbf{76.9}} & - & - & \colorbox{orange!25}{{73.5}} & - & - & - & - & - & - & - \\
\colorbox{cyan!15}{\textbf{\texttt{BTP(Ours)}}}  & 61.4$\pm$2.0 & 65.1$\pm$1.5 & \colorbox{orange!25}{{71.3$\pm$0.7}} & \colorbox{red!25}{\textbf{55.5$\pm$2.6}} & \colorbox{red!25}{\textbf{84.5$\pm$2.0}} & \colorbox{red!25}{\textbf{17.0$\pm$2.0}} & 65.2$\pm$1.1 & \colorbox{red!25}{\textbf{71.4$\pm$1.1}} & \colorbox{red!25}{\textbf{76.3$\pm$0.7}} & \colorbox{red!25}{\textbf{55.8$\pm$2.6}} & \colorbox{red!25}{\textbf{87.3$\pm$1.5}} & \colorbox{red!25}{\textbf{27.8$\pm$2.4}} \\

\bottomrule
\end{tabular}
\end{adjustbox}
\end{table*}

\begin{table*}[t]
\centering
\caption{Comprehensive ablation studies on proposed modules and training objectives.
Object-level results are reported as AUROC / AP (\%), and point-level results as AUROC / PRO (\%).}
\label{tab:ablation_all}
\footnotesize
\setlength{\tabcolsep}{9pt}
\renewcommand{\arraystretch}{1.05}
\begin{tabular}{l|ccccc|cc|cc}
\toprule
\multicolumn{6}{c|}{} & \multicolumn{2}{c|}{Real3D-AD} & \multicolumn{2}{c}{Anomaly-ShapeNet} \\
\cmidrule(lr){7-8} \cmidrule(l){9-10}
Type & MGFEM & GFCM & $\mathcal{L}_{\text{global}}$ & $\mathcal{L}_{\text{local}}$ & $\mathcal{L}_{\text{geo}}$ 
& object-level & point-level & object-level & point-level \\
\midrule
\multicolumn{10}{l}{\textbf{Ablation on proposed modules}} \\
\midrule
Baseline &  &  &  &  &  & (51.5, 54.5) & (-, -) & (53.9, 60.7) & (-, -) \\
MGFEM & \cmark &  & \cmark & \cmark &  & (59.9, 63.3) & (83.3, 54.6) & (61.4, 68.9) & (83.5, 53.8) \\
GFCM &  & \cmark & \cmark &  & \cmark & (52.5, 57.4) & (55.0, 18.7) & (53.8,61.1) & (52.0,14.5) \\
\rowcolor{gray!20}
\textbf{Full} & \cmark & \cmark & \cmark & \cmark & \cmark & \textbf{(61.4, 65.1)} & \textbf{(84.5, 55.5)} & \textbf{(65.2, 71.4)} & \textbf{(87.3, 55.8)} \\
\midrule
\multicolumn{10}{l}{\textbf{Ablation on training objectives}} \\
\midrule
\rowcolor{gray!20}
\textbf{Full} & \cmark & \cmark & \cmark & \cmark & \cmark & \textbf{(61.4, 65.1)} & \textbf{(84.5, 55.5)} & \textbf{(65.2, 71.4)} & \textbf{(87.3, 55.8)} \\
w/o $\mathcal{L}_{\text{geo}}$    & \cmark & \cmark & \cmark & \cmark &  & (61.3, 64.8) & (82.2, 50.2) & (60.9, 68.6) & (74.1, 50.0) \\
w/o $\mathcal{L}_{\text{local}}$  & \cmark & \cmark & \cmark &  & \cmark & (60.3, 64.1) & (68.8, 26.1) & (61.1, 68.8) & (62.4, 20.3) \\
w/o $\mathcal{L}_{\text{global}}$ & \cmark & \cmark &  & \cmark & \cmark & (60.5, 63.8) & (84.3, 59.1) & (60.1, 67.8) & (83.6, 57.3) \\
\bottomrule
\end{tabular}
\end{table*}

\subsection{Main Results}

\noindent\textbf{Main Evaluation.}
\cref{table:main} summarizes results on Real3D-AD under the one-class training protocol, where each model is trained on annotated data from a single category and evaluated on the remaining categories with averaged performance.
Following standard practice, we report AUROC at both the object-level (O.) and point-level (L.).
Although BTP achieves a lower object-level AUROC (61.4\%) than PointAD (74.8\%), it remains competitive in several categories (\textit{e.g. toffees}) and demonstrates superior point-level localization.
In contrast, BTP achieves the best point-level performance with a mean AUROC of 84.5\%, surpassing the second-best CPMF (75.9\%) by +8.6 points.
It ranks first in 8 out of 12 categories and second in 2 others, demonstrating strong capability for fine-grained localization where local cues dominate.
Among ZS methods (\textit{e.g.}, PointCLIPV2, AnomalyCLIP, and PointAD), BTP consistently ranks first at the point-level and second at the object-level, outperforming other ZS models and even several supervised baselines.
These results validate that the proposed multi-granularity fusion effectively enhances ZS 3D anomaly localization. For completeness, the detailed per-category results on the Anomaly-ShapeNet dataset are provided in the supplementary material.


\noindent{\textbf{Training set.} ~\cref{table:train_real3d} reports the cross-category anomaly detection results (P-AUROC) on Real3D-AD, where the model is trained on one category and evaluated across all others. The last column summarizes the average performance under each training setting. Overall, our method generalizes well, with most training categories achieving average P-AUROC above 80\%. Training on \textit{gemstone} yields the highest mean score (86.7\%), while \textit{toffees} performs relatively weaker (73.5\%), indicating that structural diversity affects transferability. Notably, even with large inter-category differences, the framework maintains strong performance (\textit{e.g.}, training on \textit{diamond} still achieves over 90\% P-AUROC on \textit{fish} and \textit{gemstone}), demonstrating robust cross-category generalization.}

\noindent{\textbf{Comprehensive Benchmarking.}
\cref{tab:all_results} reports a comprehensive comparison on Real3D-AD and Anomaly-ShapeNet across multiple metrics.
On Real3D-AD, while PointAD achieves strong object-level performance, our BTP clearly leads the point-level evaluation with 84.5\% P-AUROC and 55.5\% P-PRO.
Among ZS methods, BTP shows a large margin over PointCLIPV2, AnomalyCLIP, and PointAD across all metrics.
On Anomaly-ShapeNet, BTP maintains superior generalization, achieving the best point-level results (87.3\% P-AUROC and 55.8\% P-PRO) and competitive object-level scores.
These results demonstrate that BTP not only excels in the ZS setting but also delivers fine-grained localization performance on par with or surpassing supervised and unsupervised counterparts.}

\begin{figure}[t]
    \centering
    \includegraphics[width=0.99\linewidth]{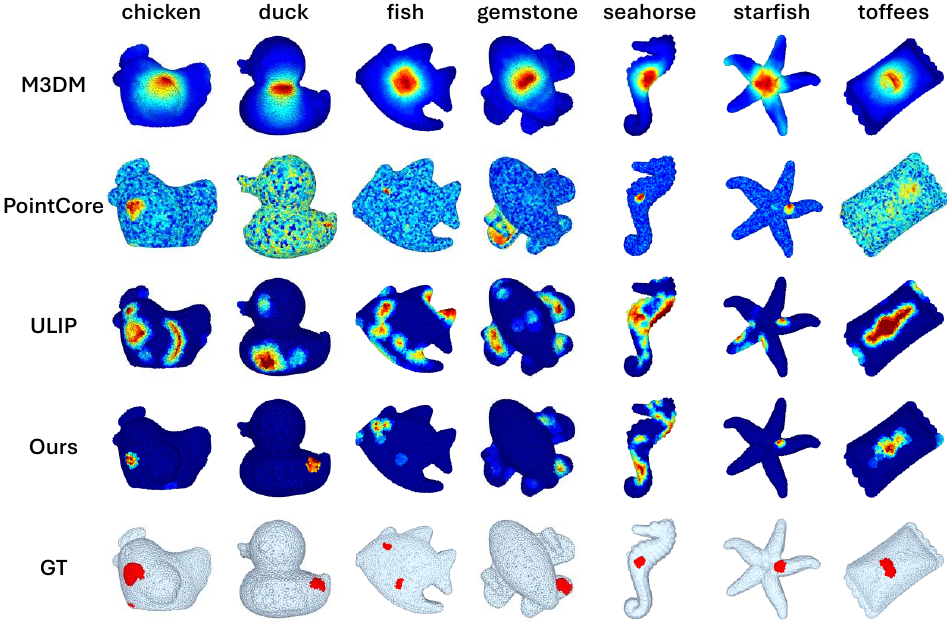}
    \caption{Visualization of anomaly localization on Real3D-AD.}
    \label{fig:main_visual}
\end{figure}

\noindent{\textbf{Qualitative Results.}~\cref{fig:main_visual} compares anomaly localization on Real3D-AD. For fairness, we equip ULIP with a simple dimensional alignment to enable point-wise scoring, since the original ULIP does not support localization. We also include M3DM \cite{wang2023multimodal} and PointCore \cite{zhao2025pointcore} as representative baselines. Our method localizes anomalous regions more accurately, highlighting its ability to capture fine-grained structural defects. More visualizations and failure cases are provided in the supplementary material.}


\subsection{Ablation Study}

\noindent\textbf{Module Ablation.}
We analyze the contribution of each proposed module on Real3D-AD and Anomaly-ShapeNet, as summarized in~\cref{tab:ablation_all}.
The baseline trained without fusion performs poorly due to the absence of semantic–geometric interaction.
Introducing MGFEM strengthens representation by capturing hierarchical point-level cues, leading to more accurate localization.
Using GFCM alone yields marginal improvement, indicating that geometric priors without semantic hierarchy are insufficient.
When both modules are combined, the model reaches the best performance (61.4\%/65.1\% object-level and 84.5\%/55.5\% point-level on Real3D-AD), confirming that MGFEM and GFCM complement to enhance geometry-aware representation.

\noindent\textbf{Training Objective Ablation.}
We further examine the effect of each training objective, as reported in~\cref{tab:ablation_all}.
Removing the global supervision $\mathcal{L}_{\text{global}}$ weakens semantic alignment between visual and geometric embeddings.
Discarding the local supervision $\mathcal{L}_{\text{local}}$ degrades fine-grained correspondence and harms point-level localization.
Without the geometric loss $\mathcal{L}_{\text{geo}}$, the model loses structural regularization and becomes less robust.
The full configuration achieves the best results (61.4\%/65.1\% object-level and 84.5\%/55.5\% point-level on Real3D-AD), verifying that all objectives collaboratively improve both semantic coherence and geometric sensitivity in ZS 3D anomaly detection.

\begin{table}[t]
\centering
\caption{Ablation study on number of input points.}
\label{tab:ablation_point}
\footnotesize
\renewcommand{\arraystretch}{1.05}
\begin{adjustbox}{max width=\linewidth}
\begin{tabular}{c|cc|cc|c}
\toprule
\#Points & O-AUROC & O-AP & P-AUROC & P-AUPRO & FPS\\
\midrule
1024  & \colorbox{orange!25}{{62.7}} & \colorbox{orange!25}{{64.9}} & \colorbox{orange!25}{{83.0}} & 52.9 & \colorbox{red!25}{\textbf{75.3}}\\
2048  & 61.4 & \colorbox{red!25}{\textbf{65.1}} & \colorbox{red!25}{\textbf{84.5}} & \colorbox{red!25}{\textbf{55.5}} & \colorbox{orange!25}{{73.8}}\\
4096  & \colorbox{red!25}{\textbf{62.9}} & 64.2 & 82.3 & \colorbox{orange!25}{{54.1}} & 71.9\\
8192  & 61.5 & 64.7 & 80.1 & 53.7 & 69.8\\
\bottomrule
\end{tabular}
\end{adjustbox}
\end{table}

\noindent{\textbf{Point Number Ablation.}
\cref{tab:ablation_point} evaluates different input point numbers on Real3D-AD.
Point-level metrics peak at 2048 points (P-AUROC 84.5\%, P-AUPRO 55.5\%), indicating the best localization.
While 4096 points slightly improves O-AUROC, the gain is marginal and speed drops from 75.3 FPS (1024) to 69.8 FPS (8192).
Balancing accuracy and efficiency, we adopt 2048 points as the default setting, supporting real-time industrial inspection.}

\section{Conclusion}
\label{sec:con}
In this paper, we proposed BTP (Back To Point), a ZS anomaly detection framework that operates directly in the intrinsic 3D geometric domain with pre-trained point-language models (PLMs). Unlike image-based pipelines that rely on 2D projection and reprojection, BTP performs point-centric feature learning to establish intrinsic alignment between geometric and linguistic representations, preserving fine-grained spatial topology for accurate localization of subtle defects. Through multi-granularity feature alignment and joint representation learning, BTP effectively integrates global semantics with local geometric cues, enhancing the discriminative power and robustness of cross-modal embeddings. Extensive experiments on Real3D-AD and Anomaly-ShapeNet demonstrate that BTP consistently improves both object-level detection and point-level localization, while also revealing the potential of PLMs to unify 3D geometric reasoning and language understanding for industrial inspection. In future work, we will explore more effective object-level scoring and aggregation strategies to further close the gap between fine-grained localization and global anomaly recognition.
\section*{Acknowledgment}
This work was supported by the National Natural Science Foundation of China (62576182, 62401305), Young Talent of Lifting engineering for Science and Technology in Shandong, China (SDAST2025QTA004), and the Qilu Youth Innovation Team (2024KJH028).

{
    \small
    \bibliographystyle{ieeenat_fullname}
    \bibliography{ref}
}


\end{document}